\documentclass[11pt,a4paper]{article}

\usepackage[margin=1in]{geometry}
\usepackage[T1]{fontenc}
\usepackage[utf8]{inputenc}
\usepackage{lmodern}
\usepackage{microtype}

\usepackage[colorlinks,allcolors=blue!60!black]{hyperref}

\usepackage{amsmath,amsthm,amssymb}
\usepackage{mathtools}
\usepackage[authoryear,longnamesfirst]{natbib}
\usepackage{cleveref}
\usepackage{algorithm}
\usepackage{algpseudocode}

\usepackage{tikz}

\usepackage{authblk}

\usepackage{thmtools}
\usepackage{thm-restate}

\theoremstyle{definition}

\newtheorem{q}{Question}

\DeclareMathOperator{\loss}{loss}
\newcommand{\CCC}{\ensuremath{\mathcal C}}

\title{The Core in Max-Loss Non-Centroid Clustering Can Be Empty }

\author[1]{Robert Bredereck}
\author[1]{Eva Deltl}
\author[1]{Leon Kellerhals}
\author[2]{Jannik Peters}

\affil[1]{TU Clausthal, Germany}
\affil[2]{National University of Singapore, Singapore}

\date{} 

\begin{document}
\maketitle

\begin{abstract}
We study core stability in non-centroid clustering under the max-loss objective, where each agent’s loss is the maximum distance to other members of their cluster. We prove that for all $k\geq3$ there exist metric instances with $n\ge 9$ agents, with $n$ divisible by $k$, for which no clustering lies in the $\alpha$-core for any $\alpha<2^{\frac{1}{5}}\sim 1.148$. The bound is tight for our construction. Using a computer-aided proof, we also identify a two-dimensional Euclidean point set whose associated lower bound is slightly smaller than that of our general construction. This is, to our knowledge, the first impossibility result showing that the core can be empty in non-centroid clustering under the max-loss objective.
\end{abstract}



\section{Introduction}\label{intro}
Clustering is a fundamental task in data analysis, optimization, and beyond: we are given $n$ points (or agents) with the goal to partition these points into $k$ groups, the so-called \emph{clusters}. Most works in clustering focus on the \emph{centroid model}. In this model, each cluster has a representative center. For a given point, the \emph{loss} of this point is its distance to the nearest center or equivalently to the center assigned to its partition.
Recent work in clustering has introduced and studied many variations of \emph{proportional fairness} \citep{DBLP:journals/corr/abs-1905-03674,micha_et_al:LIPIcs.ICALP.2020.85,ALMV24,KePe24},
where a clustering is considered proportional if there is no coalition of at least $\tfrac{n}{k}$ agents that can agree on a new representative which strictly reduces the loss of each coalition member.

The above-mentioned work solely focuses on centroid clustering which always selects a representative for each of the clusters.
However, there are also settings where clusters do not have natural representatives, for example in settings of team formation \citep{coalitionForm2,woeginger2013coalition}
or clustered federated learning \citep{sattler2021clustered}.
In these \emph{non-centroid} (representative-free) settings, an agent's loss depends only on distances to other members of its cluster.
\cite{CMS24} extended proportional fairness to this setting: A (non-centroid) clustering is in the \emph{$\alpha$-core} if no coalition of at least $\tfrac{n}{k}$ agents can form a new cluster in which each member's loss decreases by a factor of more than $\alpha$.

Among other possible definitions of loss, \citet{CMS24} consider the \emph{max-loss objective}, where each agent's loss in a cluster is the maximum distance to another agent in the cluster.
For this objective, they show that there always exists a clustering which is in the $2$-core. 
Moreover, they state the following open problem.
\begin{q}[\citet{CMS24}]
    For the max-loss objective, does there always exist a clustering in the $1$-core?
\end{q}
The same question was repeated by \citet[page~8]{CSY25a}(who studied how to combine both proportional centroid and non-centroid clustering).

As our main contribution, we answer the question in the negative.

\begin{restatable}{theorem}{nope}
\label{thm:nope}
For every $k \ge 3$ there exists a metric space with $n \ge 9$ agents, where $n$ is divisible by~$k$, for which no $k$-clustering is in the $\alpha$-core under the max-loss objective, for any $\alpha < 2^{\frac{1}{5}} < 1.149$.
\end{restatable} 

The metric space we use to prove \Cref{thm:nope} is not Euclidean, whereas most metric spaces considered in the clustering setting are.
Using a computer-aided proof, we were able to find a set of points in two-dimensional Euclidean space which yield a slightly smaller lower bound.
These points were obtained by optimizing a system of inequalities and are based on the metric space used for proving \Cref{thm:nope}.

\begin{restatable}{theorem}{euclnope}
	\label{thm:euclnope}
	For 
$k = 3$ there is a two-dimensional Euclidean space 
with $n = 9$ many agents for which no $k$-clustering is in the $\alpha$-core under the max-loss objective, for any $\alpha < 1.138$.
\end{restatable}

The points for the counterexample, along with a Python script for verifying the claim, can be found in Appendix \ref{app:euclnope}.

\paragraph{Notation.}
Let $(N, d)$ be a (pseudo)-metric space with a distance function $d \colon N \times N \to \mathbb R_{\ge 0}$ satisfying
$d(i,i) = 0$, $d(i,j) = d(j, i)$ and $d(i, j) + d(j, h) \ge d(i, h)$ for all $i, j, h \in N$.
We call a metric space $(N,  d)$ $t$-dimensional Euclidean if $N \subseteq \mathbb R^t$ and if $d$ is the Euclidean distance function, i.e., $d(i,j) = \sqrt{(i-j)^2}$.

A (non-centroid) $k$-clustering of a metric space $(N, d)$ of $n$ agents is a partition $\mathcal{C}=(C_1,\dots,C_k)$ of $N$ into $k$ (nonempty) clusters.
Denote by $\mathcal{C}(i)$ the cluster including $i$.
For $i\in N$ and a set $S\subseteq N$, the \emph{individual max-loss} of $i$ in $S$ is
\[
\loss_i(S)\coloneqq \max_{y\in S} d(i,y).
\] 
For a clustering $\mathcal{C}$, we write $\loss_i(\mathcal{C})\coloneq \loss_i(\mathcal{C}(i))$.

We say that a coalition $S \subseteq N$ of agents $\alpha$-blocks a clustering $\CCC$ if $\alpha \cdot \loss_i(S) < \loss_i(\CCC)$ for all $i \in S$.
A $k$-clustering $\CCC$ is in the $\alpha$-core for $\alpha \ge 1$ if there is no $\alpha$-blocking coalition of size at least $\frac{n}{k}$.
We refer to the $1$-core simply as the core.

\section{Proof of \Cref{thm:nope}}\label{results}

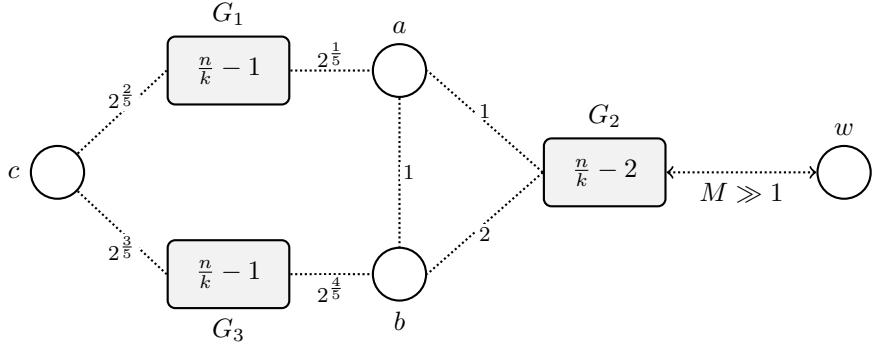
\begin{figure}
        \centering
\begin{tikzpicture}[
scale=0.9,
  every node/.style={font=\small},
  hub/.style={circle,draw,thick,minimum size=7mm,inner sep=0pt},
  grp/.style={rectangle,draw,thick,rounded corners=1mm,minimum width=16mm,minimum height=9mm,inner sep=2pt,fill=gray!10},
  meas/.style={densely dotted,thick},
  w/.style={midway,fill=white,inner sep=1pt,font=\scriptsize}
]

\node[hub,label=left:{$c$}]  (C) at (-4.5,0) {};

\node[grp,label=above:{$G_1$}] (G1) at (-2,  1.5) {$\tfrac{n}{k}-1$};
\node[grp,label=below:{$G_3$}] (G3) at (-2, -1.5) {$\tfrac{n}{k}-1$};

\node[hub,label=above:{$a$}] (A) at (0.5, 1.5) {};
\node[hub,label=below:{$b$}] (B) at (0.5,-1.5) {};

\node[grp, label=above:{$G_2$}] (G2) at (3.5,0) {$\tfrac{n}{k}-2$};

\node[hub,label=above:{$w$}] (D) at (7,0) {};

\draw[meas] (C) -- (G1.west) node[w,above] {$2^{\frac{2}{5}}$};
\draw[meas] (C) -- (G3.west) node[w,below] {$2^{\frac{3}{5}}$};

\draw[meas] (G1.east) -- (A.west) node[w,above] {$2^{\frac{1}{5}}$};
\draw[meas] (G2.west) -- (A.east) node[w,above] {$1$};

\draw[meas] (G2.west) -- (B.east) node[w,below] {$2$};
\draw[meas] (G3.east) -- (B.west) node[w,below] {$2^{\frac{4}{5}}$};

\draw[meas] (A) -- (B) node[w,right] {$1$};

\draw[<->,densely dotted,thick] (G2.east) -- (D.west)
  node[midway,below] {$M \gg 1$};

\end{tikzpicture}

        \caption[Counterexample]{Counterexample for non-centroid max-loss: edge labels specify their length; the distance between any two points is the length of the shortest path.}
\label{fig:nomaxcore}
    \end{figure}

Fix $k=3$ and let $n \ge 9$ divisible by $3$.
We create three sets of agents, $G_1$, $G_2$, and $G_3$, which we will henceforth call \emph{groups}, with
\[
|G_1|=|G_3|=\frac{n}{k}-1,\; |G_2|=\frac{n}{k}-2,
\]
and
$d(i,j)=0$ for each $i, j \in G_t$ and each $t \in \{1,2,3\}$.
The remaining four agents will be called $a, b, c$, and $w$.
We have the following distances for all agent--group pairs (as illustrated in Figure~\ref{fig:nomaxcore}).
\[
\begin{array}{c|ccc}
& G_1 & G_2 & G_3\\\hline
a & 2^{\frac{1}{5}}  & 1 & (1 + 2^{\frac{4}{5}})  \\
b & (1+2^{\frac{1}{5}})  & 2  & 2^{\frac{4}{5}} \\
c & 2^{\frac{2}{5}}  & (1+2^{\frac{3}{5}}) & 2^{\frac{3}{5}} 
\end{array}
\]
Moreover, we have $d(a, b) = 1$,
and agent $w$ is very far away from all other agents, that is, $d(w, i) = M$ for all $i \ne w $ and a sufficiently large $M \gg 1$.
The remaining undefined distances can be chosen as the lengths of shortest paths on the weighted graph induced by the specified edges.

In our proof,
we focus on the five following coalitions of size $\tfrac{n}{k}$:
\begin{align*}
S_1\coloneq G_1\cup &\,\{a\},\quad
S_2\coloneq G_1\cup\{c\},\quad
S_3\coloneq G_2\cup\{a,b\},\\
&S_4\coloneq G_3\cup\{b\},\quad
S_5\coloneq G_3\cup\{c\}.
\end{align*}

For the sets $S_1$, $S_2$, $S_4$ and $S_5$, and each agent $i$ within that set, we have

\begin{align*}
\loss_i(S_1)=2^{\frac{1}{5}},\;
\loss_i(S_2)=2^{\frac{2}{5}},\;
\loss_i(S_4)=2^{\frac{4}{5}},\;\;  \text{and} \;\;
\loss_i(S_5)=2^{\frac{3}{5}}.
\end{align*}
For $S_3$ and each $i \in S_3 \setminus \{a\}$ we have $\loss_i(S_3) = 2$ while the loss of agent $a$ is $\loss_a(S_3)=1$.

Let $\mathcal{C}=(C_1,C_2,C_3)$ be any $3$-clustering of the gadget.
One cluster must contain $w$, and any such cluster has a loss of $M$ for all its members, unless $w$ is placed alone as a singleton. Observe that the cluster containing $w$ contains at most $\tfrac{n}{k}$ members, otherwise the cluster's agents without $w$ form a blocking coalition for an $\alpha$-core, with $\alpha\geq\frac{M}{2}$.

We distinguish between two cases,
namely whether or not $\mathcal C$ contains a \emph{mixed cluster}.
Here, we call a cluster in $\mathcal C$ mixed if it contains agents from more than one of the three groups $G_1, G_2, G_3$.


\medskip
\noindent\textbf{Case 1: There is no mixed cluster.}
We claim that, if $\CCC$ has a separate cluster for each of the groups $G_1, G_2, G_3$, then one of $S_1, \dots, S_5$ is an $\alpha$-blocking coalition.

If each group lies in its own cluster, then one of the three must also contain $w$.
As mentioned above, we assume without loss of generality the cluster containing $w$ has size at most $\frac{n}{k}$.
\begin{itemize}
\item \emph{$G_1$ with $w$.}
	Then coalition $S_2 = G_1 \cup \{c\}$ $\alpha$-blocks $\CCC$:
	Clearly, the loss of each agent $i \in G_1$ in $S_2$ is smaller than in $\CCC$ by a factor greater than $\alpha$.
	Since $|\mathcal{C}(w)| \leq \frac{n}{k}$, agent $c$ must be clustered with agents from $G_2$ or $G_3$;
	thus its loss is
	\[ \loss_c(\CCC) \ge 2^{\frac{3}{5}} > \alpha 2^{\frac{2}{5}} = \alpha \loss_c(S_2). \]
\item \emph{$G_3$ with $w$.}
	Suppose first that $b$ is clustered with $G_1$.
	Then due to our above assumption of having no mixed cluster, the third cluster contains $G_2$. 
    For $c$, this implies that it is clustered either with $b$ or with $G_2$, thus
	\[ \loss_c(\CCC) \ge 2^{\frac{3}{5}} > \alpha 2^{\frac{2}{5}} = \alpha \loss_c(S_2). \]
    And for all $i \in G_1$ \[ \loss_i(\CCC) \ge d(b,i) > 2^{\frac{3}{5}} > \alpha 2^{\frac{2}{5}} = \alpha \loss_i(S_2). \] So $S_2 = G_1 \cup \{c\}$ $\alpha$-blocks $\CCC$.
    
	If $b$ is clustered with $G_2$, then we claim that $S_4 = G_3 \cup \{b\}$ $\alpha$-blocks $\CCC$.
	Again this is obvious for the members of $G_3$, and for $b$, we have
	\[
		\loss_b(\CCC) \ge 2 > \alpha 2^{\frac{4}{5}} = \alpha \loss_b(S_4).
	\]
\item \emph{$G_2$ with $w$.} 
\begin{itemize}
    \item
	    If $b$ is clustered with $G_2$ and $w$, then $S_3=G_2\cup\{a,b\}$ $\alpha$-blocks $\CCC$.
	    The improvement is clear for all agents clustered with $w$.
	    For agent $a$ we have
	    $\loss_a(\CCC) \ge 2^{\frac{1}{5}} > \alpha = \alpha \loss_a(S_3).$
    \item If $b$ is clustered with $G_1$, leading to a loss of at least $(1+2^{\frac{1}{5}})$ for all its members, then $S_2 = G_1\cup \{c\}$ $\alpha$-blocks, as $\loss_c(\CCC) \ge 2^{\frac{3}{5}} > \alpha 2^{\frac{2}{5}} = \alpha \loss_c(S_2).$
    \item Let $b$ be clustered with $G_3$.
	    If $a$ and $c$ are clustered with $G_1$, then $S_1 = G_1 \cup \{a\}$ $\alpha$-blocks as $\loss_i(\CCC) \ge 2^{\frac{2}{5}} > \alpha \cdot 2^{\frac{1}{5}} = \loss_i(S_1)$ for all $i \in S_1$.
	    Otherwise, one of $a$ and $c$ is not clustered with $G_1$.
	    If $c$ is with $G_3 \cup \{b\}$ or with $G_2 \cup \{w\}$,
	    then $S_5 = G_3 \cup \{c\}$ $\alpha$-blocks $\CCC$ as
	    $\loss_i(\CCC) \ge 2^{\frac{4}{5}} > \alpha \cdot 2^{\frac{3}{5}} = \alpha\loss_i(S_5)$ for all $i \in S_5$.
	    If instead $a$ is clustered with $G_3 \cup \{b\}$ or with $G_2 \cup \{w\}$,
	    then $S_1 = G_1 \cup \{a\}$ $\alpha$-blocks $\CCC$ as
	    $\loss_i(\CCC) \ge 2^{\frac{2}{5}} > \alpha \cdot 2^{\frac{1}{5}} = \alpha\loss_i(S_1)$ for all $i \in S_1$.
\end{itemize}
\end{itemize}
In all cases, we find an $\alpha$-blocking coalition for some $\alpha \ge 2^{\frac{1}{5}}$.

\paragraph{Case 2: There is a mixed cluster.}

We claim that one of $S_1, \dots, S_4$ $\alpha$-blocks $\CCC$.

Suppose that there is only one mixed cluster and call it $C_1$.
As the other two clusters are not mixed, $C_1$ must contain all members of one of the groups.
Assume the mixed cluster contains~$w$.
Since the cluster containing $w$ cannot have more than $\frac{n}{k}$ agents, $C_1$ must contain the agents of~$G_2$, and one other agent $x$ from $G_1$ or $G_3$.
Now consider the clustering $\CCC'$ where~$x$ is moved to the cluster containing the remaining members of its group.
Then $\CCC'$ does not contain a mixed cluster and therefore, by Case 1, it is not in the $\alpha$-core.
Now observe that for every agent $i \in N$ we have $\loss_i(\CCC') \le \loss_i(\CCC)$;
thus if a coalition $S$ $\alpha$-blocks $\CCC'$, then it also $\alpha$-blocks $\CCC$.
Consequently, $\CCC$ is also not in the $\alpha$-core.

Hence, we may assume our clustering $\CCC$ contains a mixed cluster that does not contain $w$.
In any such clustering $\CCC$, every agent in a mixed cluster has loss at least $1+2^{\frac{1}{5}}$, and every agent in the cluster containing $w$ has loss at least $M\gg (1+2^{\frac{1}{5}})$.
If there is an agent with a smaller loss, then it lies in the third cluster, which then is not mixed.
It follows that there are at least two groups whose members all incur a loss of at least $1 + 2^{\frac{1}{5}}$ in $\CCC$: One group in the cluster containing $w$ and at least one group in the mixed cluster.
We call such groups \emph{deviating}.

\begin{itemize}
    \item Suppose that $G_1$ is deviating.
If $a$ and $c$ are part of the same cluster, then $S_1 = G_1 \cup \{a\}$ is $\alpha$-blocking:
For each $i \in G_1$, we have
\[ \alpha \cdot \loss_i(S_1) = \alpha \cdot 2^{\frac{1}{5}} < 2^{\frac{2}{5}} < 1+ 2^{\frac{1}{5}} \leq \loss_i(\CCC),\]
and for $a$ we have 
\[ \alpha \cdot \loss_a(S_1) = \alpha \cdot 2^{\frac{1}{5}} < 2^{\frac{2}{5}} < 2^{\frac{1}{5}} + 2^{\frac{2}{5}} = d(a, c) \le \loss_a(\CCC). \]
Similarly, if $b$ and $c$ are part of the same cluster,
then $S_2 = G_1 \cup \{c\}$ is $\alpha$-blocking as for each $i \in G_1$,
\[ \alpha \cdot \loss_i(S_2) = \alpha \cdot 2^{\frac{2}{5}} < 2^{\frac{1}{5}} \cdot 2^{\frac{2}{5}} \le 1+2^{\frac{1}{5}} = \loss_i(\CCC), \]
and for $c$ we have
\[ \alpha \cdot \loss_c(S_2) = \alpha \cdot 2^{\frac{2}{5}} < 2^{\frac{3}{5}} + 2^{\frac{4}{5}} = d(b, c). \]
Finally, if $a$ or $c$ are clustered with $w$ or with $G_3$, then again $S_1$ or $S_2$ $\alpha$-blocks, because both $a$ and $c$ face losses exceeding $2^{\frac{1}{5}}$ or $2^{\frac{2}{5}}$, respectively, i.e., by a factor strictly larger than $2^{\frac{1}{5}}$.

\item If $G_1$ is not among the deviating groups, then $b$ cannot be clustered with $G_1$ (otherwise $G_1$ would deviate, as $d(b, G_1) = 1 + 2^{\frac{1}{5}}$). Hence $b$ is either clustered with $w$ or in a mixed cluster, and in both cases its loss exceeds $2$.
Thus $S_4 = G_3 \cup \{b\}$ is $\alpha$-blocking:
For each $i \in S_4$, we have
\[ \alpha \cdot \loss_i(S_1) = \alpha \cdot 2^{\frac{4}{5}} < 2 \leq \loss_i(\CCC).\]
\end{itemize}

Thus in every clustering where at least one cluster is mixed, one of the coalitions $S_1,\dots,S_4$ strictly blocks by a factor of at least $2^{\frac{1}{5}}$, contradicting $\alpha$-core stability.

\smallskip
Finally, to extend to arbitrary $k\ge 3$ and $n$ divisible by $k$, add $(k-3)$ dummy clusters of size $\tfrac{n}{k}$ each, with intra-distance $0$ and distance $M$ to every outside node. In any feasible $k$-clustering, these dummy clusters must stand alone, so the effective choice reduces to clustering the gadget $\{a,b,c,w,G_1,G_2,G_3\}$.

\section{Outlook}\label{outlook}
A couple of questions emerge from the two results.
Can we improve upon the upper bound of $\alpha = 2$ for the core guarantee?
Also, can we possibly prove a better core guarantee for Euclidean spaces than for metric spaces, as is the case in proportional (centroid) clustering setting \citep{DBLP:journals/corr/abs-1905-03674,micha_et_al:LIPIcs.ICALP.2020.85}? 

\bibliographystyle{abbrvnat}
\bibliography{cas-refs} 

\newpage

\appendix

\section{Points for \Cref{thm:euclnope}}
\label{app:euclnope}

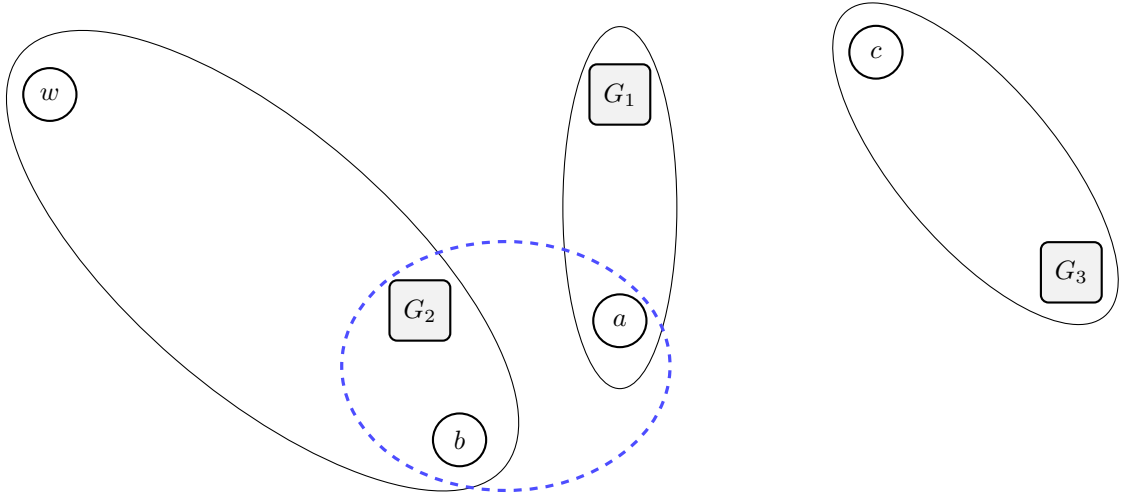
\begin{figure}
        \centering
\begin{tikzpicture}[
scale=3.0,rotate=90,
  every node/.style={font=\small},
  hub/.style={circle,draw,thick,minimum size=7mm,inner sep=0pt},
  grp/.style={rectangle,draw,thick,rounded corners=1mm,minimum width=8mm,minimum height=8mm,inner sep=2pt,fill=gray!10},
  meas/.style={densely dotted,thick},
  w/.style={midway,fill=white,inner sep=1pt,font=\scriptsize}
]

\node[hub]  (C) at ( 1.18678884,-1.12314325) {$c$};

\node[grp] (G1) at (1.0, 0.0) {$G_1$};
\node[grp] (G3) at ( 0.21399904, -1.97998602) {$G_3$};

\node[hub] (A) at (0.0, 0.0) {$a$};
\node[hub] (B) at (-0.52633831, 0.70311454) {$b$};

\node[grp] (G2) at ( 0.04545127, 0.87711815) {$G_2$};

\node[hub] (D) at (1.0,2.5) {$w$};

\draw (G2) ++(0.22,0.69) ellipse [x radius=1.4cm,y radius=0.58cm,rotate=49];

\draw (G3) ++(0.48,0.42) ellipse [x radius=0.88cm,y radius=0.35cm,rotate=40];

\draw (0.5,0.0) ellipse [x radius=0.8cm, y radius=0.25cm];

\draw[very thick, dashed, blue!70] (A) ++(-.2,0.5) ellipse [x radius=0.55cm, y radius=0.72cm];






\end{tikzpicture}

        \caption[Counterexample]{
		The points for the counterexample in two-dimensional Euclidean space (rotated by 90~degrees). $w$ was placed closer to the remaining points for an easier illustration.
		The black ellipses depict the clustering that is in the $1.139$-core and the blue (dashed) ellipse is the $1.139$-blocking coalition.
	}
\label{fig:noeuclmaxcore}
    \end{figure}

The counterexample to prove \Cref{thm:euclnope} for $n=9$ and $k=3$ uses the same set of points as in \Cref{thm:nope}, with each (group of) points having roughly the same ``function''; see \cref{fig:noeuclmaxcore} for an illustration.
The points were obtained by optimizing a system of inequalities and are as follows.

\[
\begin{aligned}
a   &: (0.0,\;0.0),\\
b   &: (-0.52633831,\; 0.70311454),\\
c   &: ( 1.18678884,\;-1.12314325),\\
G_1 &: (1.0,\;0.0),\\
G_2 &: ( 0.04545127,\; 0.87711815),\\
G_3 &: ( 0.21399904,\; -1.97998602),\\
w &: ( 10.0,\; 10.0 ).
\end{aligned}
\]

To find an $\alpha$-core clustering with the smallest $\alpha$ for the set of points, we applied \Cref{alg:corecheck} to each possible clustering $\CCC$.

\begin{algorithm}[b]
\caption{\textsc{CoreCheck}$(N, d, \CCC)$}
	\label{alg:corecheck}
\begin{algorithmic}[1]
\State $\alpha \gets 1$; $n \gets |N|$; $k \gets |\CCC|$
\For{each coalition $S \subseteq N$, $S \ge \frac{n}{k}$}
    \For{each $i \in S$}
    	\State $\rho_i \gets \loss_i(\CCC)/\loss_i(S)$
    \EndFor
    \State $\alpha \gets \max\{\alpha,\; \min_{i \in S} \rho_i\}$
\EndFor
\State \Return $\alpha$
\end{algorithmic}
\end{algorithm}

The code for the computation of the point coordinates and verification is available at:
\url{https://github.com/evamichelle30/CounterExampleMaxLossCore}.


\end{document}